\documentclass[submission,copyright,creativecommons]{eptcs}
\usepackage{graphicx,subfig}

\title{Human-Agent Decision-making: Combining Theory and Practice}

\providecommand{\event}{TARK 2015} 

\author{Sarit Kraus
\institute{Department of Computer Science\\
Bar-Ilan University\\
Ramat-Gan Israel}
\email{sarit@cs.biu.ac.il}
}
\def\titlerunning{Human-Agent Decision-making: Combining Theory and Practice}
\def\authorrunning{S. Kraus}

\begin{document}
\maketitle
\begin{abstract}
Extensive work has been conducted both in game theory and logic to model strategic interaction. 
An important question is whether we can use these theories to design agents for interacting with people? 
On the one hand, they provide a formal design specification for agent strategies. On the other hand, people do not necessarily adhere to playing in accordance with these strategies, and their 
behavior is affected by a multitude of social and psychological factors. 
In this paper we will consider the question of whether strategies implied by theories of strategic behavior can be used by automated agents that interact proficiently with people. 
We will focus on automated agents that we built that need to interact with people in two negotiation settings:
 bargaining and deliberation. For bargaining we will study game-theory based equilibrium agents and for argumentation we will discuss logic-based argumentation theory.
We will also consider security games and persuasion games and will discuss the benefits of using equilibrium based agents. 
\end{abstract}


\section{Introduction}

Agents that interact proficiently with people may be useful for training \cite{LinOK09}, 
supporting \cite{kersten2007negotiation,ElmalechSRE15,ElmalechSG15,amir2013collaborative} and even replacing people in many applications \cite{durenard2013professional,kauppi2013tools}.
We are considering the agent-human interactions as being a \emph{strategic} activity \cite{fatima2014principles}. That is,
we assume that when the automated agent engages in the interaction, it should
act as best it can to realize its preferences.
\emph{Game theory} is the mathematical
theory of strategic decision-making~\cite{maschler:2013a} and  thus it seems that
game theory might be an appropriate analytical tool for
understanding how a strategic agent can and should act, and might
also be useful in both the design of automated agents and protocols
for the interactions. However, game theory assumes that all players will act as best they can to
realize their preferences. Unfortunately, humans tend to
make mistakes, and they are affected by cognitive, social and cultural
factors  \cite{Bazerman1992,Lax1992,ariely:2008a}.
In particular, people's observed behavior does not
correspond to game theory-based equilibrium strategies \cite{erev_predicting_1998,McKelvey92}.

Another approach for the development of automated agents is the (non-classical)-logic approach. The agent is
given  a logical representation of its environment and its desired
goals, and it reasons logically in order to generate its activities. When interacting with people, the environment consists also of the human model.
Yet, modeling 
people's behavior is a big challenge.
We have
incomplete information about the person's preferences, and we 
have to cope with the uncertainties inherent in human decision-making
and behavior. Human behavior is diverse, and cannot be
satisfactorily captured by a simple abstract model.
In particular, human decision-making
tends to be very noisy: a person may make different strategic
decisions in similar situations. 

In this paper we survey briefly a few of the agents that we built over the years that interact proficiently with people. 
In most of the cases, deploying only a game-theory approach or logical-based approach was not beneficial.
Heuristics and machine-learning techniques were augmented into the formal models to lead to agents that interact proficiently with people.
We will discuss three negotiation settings: multi-issue negotiations, games where the players interleave negotiations with resource exchange while attempting to satisfy their goals and argumentation settings.
Finally, we will discuss security games.
 
\section{Multi-issue negotiations}

Over the years we designed and implemented several automated agents for multi-issue negotiations. In multi-issue negotiations the players need to reach an agreement on several issues.
Each issue is associated with a set of possible values and the players need to agree on a specific value for each issue.
The negotiations can end with the negotiators signing an agreement or with one of the sides opting out of the negotiations. In addition, if the crisis
does not end within a pre-specified deadline then the status quo is implemented. Each outcome of the negotiations is associated with a utility score for both players. 
A summary of our agents is presented in Table~\ref{Multi-issue-table}.
{\small \begin{table*}
\centering
\caption{Multi-issue negotiations}
\label{Multi-issue-table}
\begin{tabular}{|l|c|l|l|l|} \hline
Settings&  Agent& Agent Properties & Scenarios& Significance \\ 
& Name & & &vs people \\ \hline
Bilateral, single-issue,  &  EQH & SPE with manually &  fishing  & One role \\ 
full information, complex   & &designed heuristics & dispute   &\\ 
actions, agreements not  & & & &\\
enforceable &&&& \\\hline
Bilateral, uncertainty,& QO-agent &  
 Qualitative decision-making& job interview& One role \\
 multi-issue&&Non-deterministic behavior& & \\ 
&&& tobacco  &\\ \cline{2-5}
&  KBAgent &  Machine learning, &job interview &Both roles\\ 
& &  qualitative decision-making& tobacco &\\
&&non-deterministic behavior & &\\ \cline{2-5}
&  NegoChat & KBagent algorithms, AAT & job interview&Both roles \\ 
&& Anchoring, NLP module &  &\\ \hline
 \end{tabular}
\end{table*}
}
\subsection{EQH agent}
The first agent, EQH, that we developed was for crisis scenarios and the setting was quite complex \cite{kraus2008resolving}. 
In addition to the message exchange in a semi-structured language, players could take actions during the negotiations and agreements were not enforceable.
In particular, opting out in a crisis is a stochastic action and thus the agents are uncertain about the result. In addition to the main issue of
the negotiation or opting out, there are various other parameters of an agent's action. These parameters influence the
utility of the negotiators from the crisis. Time plays an important role in the crisis \cite{wilkenfeld1995genie}. 
The specific scenario we used for the experimental study was 
a fishing dispute between Canada and Spain.
We formalized the crisis scenario as a game and identified a subgame-perfect equilibrium. We ran 
preliminary experiments when the automated agent followed its subgame perfect equilibrium strategy. However, the human negotiators who negotiated
with it became frustrated and the negotiation often ended with no agreement. The frustration of the human negotiators was mainly due the lack of flexibility of the agent. Since the proposed and
accepted agreements of the subgame perfect negotiation did not change over the negotiation time, the agent did not compromise.

To address this limitation of the equilibrium-based agent, we incorporated several heuristics to the EQH agent. 
We allowed the owner of the agent to determine the way the agent will deviate from the equilibrium strategies by
determining parameters that influence the agent's behavior which are instantiated before the beginning of negotiations.
In order to provide the agent with some flexibility when playing against people, we allowed the agent to consent to
agreements that have a lower utility than it would have obtained according to the relevant subgame perfect equilibrium strategy agreement.
Therefore we added the margin parameter that determines the largest number of points lower than the desired utility
value to which the agent will agree.

An additional parameter is the number of negotiation units by which the agent will increase or decrease its first
offer from the agreement specified in its equilibrium strategy. Human negotiators usually begin negotiations with an offer
higher (or lower, depending on the negotiator's role) than the value they would eventually like to reach at the end of
negotiations. This leaves bargaining space and our agent uses this type of strategy.
Another parameter indicates whether the agent will send the first message in the negotiation or will wait for its
opponent to make the first offer. The default value of this
parameter, following some literature recommendations \cite{galinsky2001first}, was that the agent will send the first offer, since we wanted a trigger to initiate negotiations with the other
agent.

Another heuristic concerns opting out. Given our assumptions, while rational agents will not opt out, people may
opt out. If the agent's expected utility from opting out is higher than its expected utility from its opponent opting out,
it will try to predict whether its opponent is going to opt out. If so, it will opt out first. The heuristic for the prediction
of whether an opponent will opt out is based on the messages sent by the opponent. For example, when a threatening
message is received, or when a comment message indicating that the negotiations are heading in a dangerous direction
is received, the estimation that the opponent may opt out increases.

We ran extensive experiments for evaluating the equilibrium agent with the heuristics (EQH agent)  \cite{kraus2008resolving}.
We compared the results of the humans to
those of the agents and concluded that the EQH agent received a higher utility score playing both roles, but the results were only statistically significant when the agent played just one of the roles. Furthermore, when an agent participates in a negotiation, the sum of the utilities are significantly higher than
when two humans play since the agent always proposes Pareto-optimal offers while people reach agreements that are not.

While the EQH agent was based on the subgame perfect equilibrium strategies, it required the introduction of many heuristics, and its success compared with people was only in one role.
The main open question is whether it is possible to provide formal methodology that will lead to an agent that is similar to the EQH without the need to manually design the EQH heuristics.
Furthermore, we are aiming for an agent that can achieve a significantly higher utility score than people in both roles. 
Toward this challenges, we next tried to use a qualtative approach, to introduce incomplete information into the environment and to improve the agent's results in both roles.
\subsection{QOagent and KBagent}

The QOagent was designed to interact with people in environments of bilateral negotiations with incomplete information when the agreements consist of multiple issues \cite{lin2008negotiating}.
With respect to incomplete information,
each negotiator keeps his preferences private, though
the preferences might be inferred from the actions of each
side (e.g., offers made or responses to offers proposed). Incomplete
information is expressed as uncertainty regarding
the utility preferences of the opponent, and it is assumed
that there is a finite set of different negotiator types. These
types are associated with different additive utility functions
(e.g., one type might have a long term orientation regarding
the final agreement, while the other type might have
a more constrained orientation). Lastly, the negotiation is
conducted once with each opponent. The experiments were run on two distinct domains. In the first domain,
England and Zimbabwe negotiate in order to reach an agreement evolving from
the World Health Organization's Framework Convention on Tobacco Control, the
world's first public health treaty. In the second domain a negotiation takes place
after a successful job interview between an employer and a job candidate.

We first formalized the scenario as a Bayesian game and computed the Bayesian Nash equilibrium.
Though we did not run simulations of the Bayesian Nash equilibrium agent against human negotiators, we ran two humans negotiations.
We found out that the opponent's utility score from the offers suggested by the equilibrium agent
are much lower than the final utility values of the human negotiations. By also
analyzing the simulation process of the human negotiations, we deduced that
without incorporating any heuristics into the equilibrium agent, the human players
would not have accepted the offers proposed by it which will lead to low utility scores for the equilibrium agent, similar to the low score of the equilibrium agent in the fishing dispute.

Therefore, we developed the \textsc{QOAgent}. For the decision-making process, the approach used by the \textsc{QOAgent}
tries to take the utility of both sides into consideration. While the
\textsc{QOAgent}'s model applies utility functions, it is based on a
non-classical decision-making method, rather than focusing on
maximizing the expected utility: the maximin function and a
qualitative valuation of offers. Using these methods, the
\textsc{QOAgent} generates offers and decides whether to accept or
reject proposals it has received.
As for incomplete information, the \textsc{QOAgent} tackles this
problem using a simple Bayesian update mechanism. After each action,
this mechanism tries to infer which negotiator type best suits the opponent. 

The effectiveness of this
method was demonstrated through extensive empirical experiments by
\cite{lin2008negotiating}.

The results of the experiments showed that the automated agent
achieved higher utility scores than the human counterpart. This can be explained
by the nature of our agent both in reference to accepting offers and generating
offers. Using the decision-making mechanism we allow the agent to propose agreements
that are good for it, but also reasonable for its opponent. In addition, the
automated agent makes straightforward calculations. It evaluates the offer based on
its attributes, and not based on its content. In addition, it also places more weight
on the fact that it loses or gains as time advances. This is not the case, however,
when analyzing the logs of the people. It seems that people put more weight on the
content of the offer than on its value. This was more evident in the Job Candidate
domain with which the human subjects could more easily identify.
Yet, this does not explain why, in both domains, similar to the EQH agent experiments, these results are significant only
for one of the sides. In the England-Zimbabwe domain, the results are significant
when the agent played the role of England, while in the Job Candidate domain these
results are significant when it played the role of the job candidate. 

In order to improve the \textsc{QOAgent}, we 
extended it by using a generic
opponent modeling mechanism, which allows the agent to model its
counterpart's population and adapt its behavior to that population \cite{oshrat:2009a}.
The extended agent, called \textsc{KBAgent}, is an automated
negotiator that negotiates with each person only once, and uses past
negotiation sessions of others as a knowledge base for generic
opponent modeling. The database containing the a relatively small
number of past negotiation sessions is used to extract the likelihood
of acceptance of proposals and which proposals may be offered by the
opposite side.  The performance of \textsc{KBAgent} in terms of its
counter-offer generation and generic opponent modeling was tested
against people in the Tobacco and the Job interview domains.

The results of these tests indicate that the
\textsc{KBAgent} negotiates proficiently with people and even achieves
higher utility score values than the \textsc{QOAgent}. Moreover, the
\textsc{KBAgent} achieves significantly better agreements, in terms of
utility score, than the human counterparts in {\it both} roles.
These results indicate that integrating general opponent
modeling into qualtative decision-making is beneficial for automated negotiations.  

\subsection{NegoChat Agent}

\begin{figure}
\centering
\includegraphics[width=3.8in]{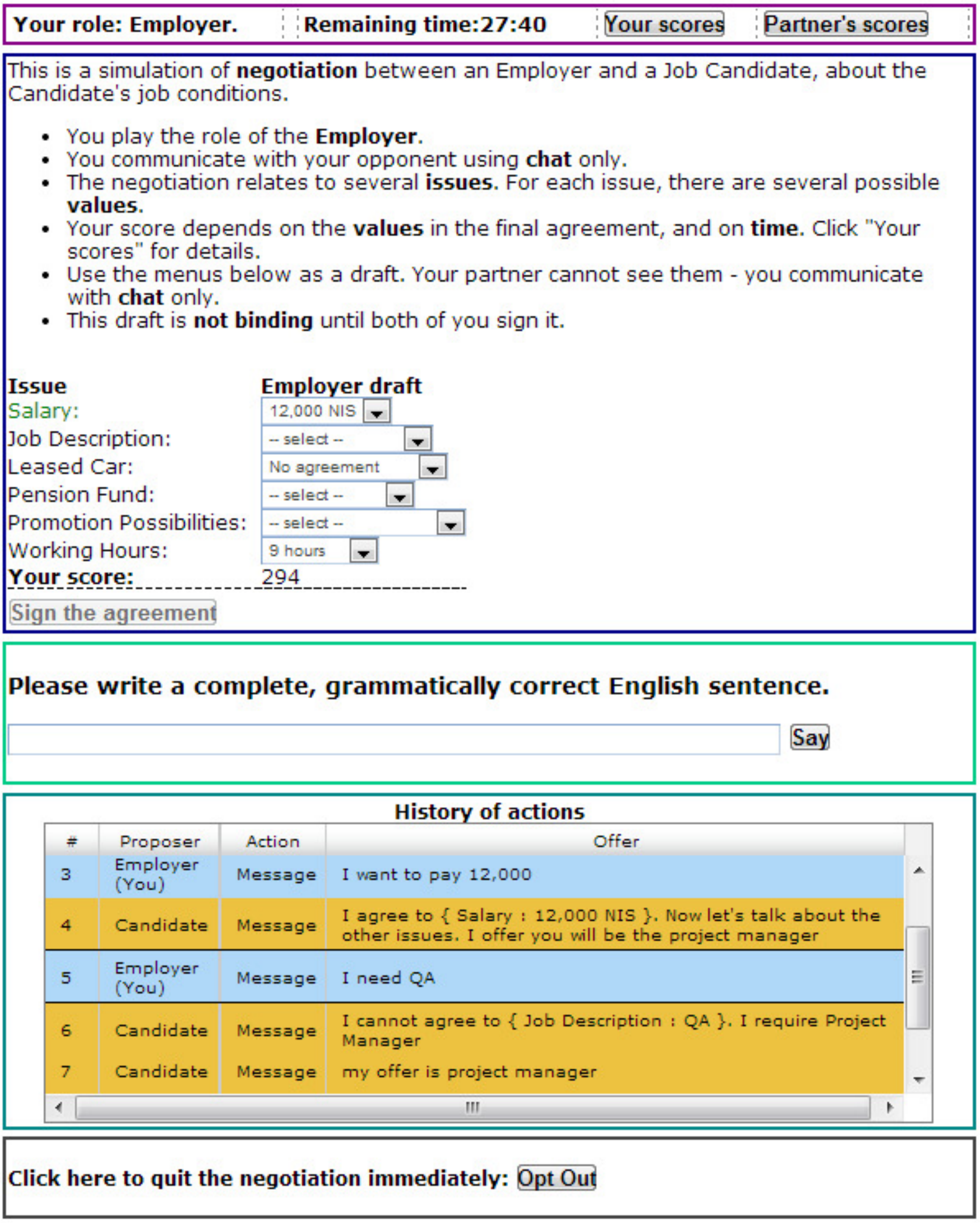}
\caption{The negotiation system's interface for NegoChat.}
\label{fig::chat}
\vspace{-5pt}
\end{figure}

All the agents we discussed so far negotiated with the human counterpart either using a structured language or using a menu-driven interaction.
They lack the natural language processing support required to enable real world
types of interactions.
To address this challenge we first developed an NLP module that translates the free text of the human player to the agent's formal language. We modified the KBagent by adding this module without changing the KBagent 
strategy
and ran an experiment in which the modified KBagent played with people in a chat-like environment (see Figure~\ref{fig::chat} for the negotiation system's interface for chat-based negotiations).
We found that simply modifying
the KBagent to include an NLP module is insufficient to create a good agent for such settings and the revised agent achieved relatively low utility scores.
The main observation was that people in chat-based negotiations make and accept partial agreements and follow 
issue-by-issue negotiations while the KBagent proposes full offers and has difficulties reaching partial agreements.
To address this limitation, we developed NegoChat, which extended the KBagent 
focusing on strategies that allow for partial agreements
and issue-by-issue interactions. 
NegoChat's algorithm is based on bounded rationality, specifically anchoring and Aspiration
Adaptation Theory (AAT). The AAT was used for deciding on the order in which the issues will be discussed. The agent begins each negotiation interaction by proposing a
full offer based on the KBagent's strategy, which serves as its anchor. Assuming this offer is not accepted, NegoChat then
proceeds to negotiate via partial agreements, proposing the next issue for negotiation based
on people's typical urgency (according to AAT).

We evaluated the NegoChat agent in extensive experiments negotiating with people in the job interview domain. We compared its performance to the performance of the 
KBAgent that also negotiated with (different) people using the same NLP module.
The NegoChat agent achieved
significantly better agreements (i.e., higher utility score) in less time. However,
people playing against KBAgent, on average, did better. This implies that some of NegoChat's
success is evidently at the cost of the person's score and consequently the social welfare
score of this agent is not significantly better than that of KBAgent. As our goal is to maximize
the agent's utility score this should not be seen as a fault. However, future generations of
automated agents may decide to implement different strategies to maximize social welfare.

\begin{figure*}[t]
\centering
\subfloat[Symmetric Board Game]{
  \includegraphics[scale=0.58]{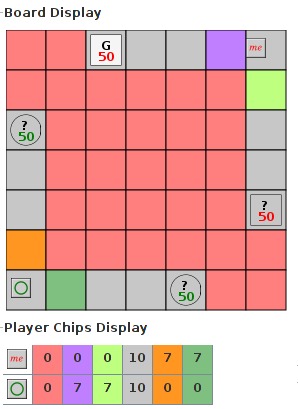}
\label{fig:sym}
}
\subfloat[Asymmetric Board Game]{
  \includegraphics[scale=0.5]{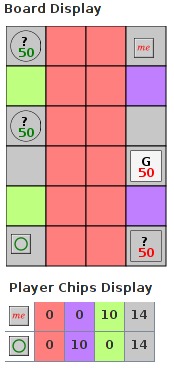} 
\label{fig:asym}
}
\subfloat[A possible proposal]{
 \includegraphics[scale=0.4]{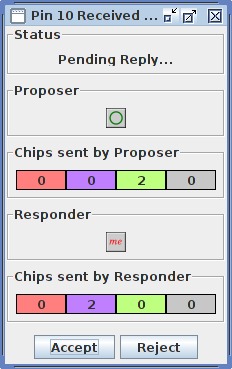}
\label{fig:prop}
}
\caption{\label{fig:ctfig} Two CT revelation games}
\end{figure*}

\section{Negotiations and Actions Interleaving}

In most situations, negotiation is not done in isolation but is associated with agreement implementation and other activities.
We developed agents that can interact with people in such settings. These studies were carried out in a configurable system called Colored
Trails (CT)\footnote{See  \texttt{http://www.eecs.harvard.edu/ai/ct}.}.  It is a game played by two or more participants on a board of
colored squares.  
CT is an abstract, conceptually simple but highly
versatile game in which players negotiate and exchange resources to
enable them to achieve their individual or group goals. It provides a
realistic analogue to multi-agent task domains, while not requiring
extensive domain modeling \cite{gal2010agent,grosz2004influence}.
A summary of the agents we developed are specified in Table~\ref{CT-table}.

 \begin{table*}
\centering
\caption{CT games }
\label{CT-table}
\begin{tabular}{|l|c|l|l|} \hline
Settings& Agent Name &Agent Properties &Significance \\ 
& & & vs people \\ \hline
Bilateral, Uncertainty,  &  PBE agent & Bayesian perfect equilibrium& No roles \\ \cline{2-4}
Revelation Game two-phases: & SIGAL & decision theory, machine learning &Both roles \\
 revelation, bargaining&&&\\ \hline
Bilateral, full information,  & PAL & decision theory: Influence Diagram &Both roles\\ 
agreements not enforceable, & & machine learning & \\
multiple rounds,  three phases:   & & & \\
bargaining, resource exchange, & & & \\ 
 movement &&&\\\hline
Contract game, three players & SPE agent & subgame-perfect equilibrium & CS role \\ \cline{2-4}
two-phases: bargaining,  & SP-RAP & subgame-perfect equilibrium & \\
movement& &bounded rational model of opponent& SP role \\ 
&& risk averse & \\ \hline
\end{tabular}
\end{table*}

\subsection{Revelation games}
\label{SIGAL-sec}

We considered negotiation settings in which participants lack information
about each other's preferences, often hindering their ability to reach beneficial agreements \cite{peled2015study}.
Specifically, we studied a particular class of such settings we call ``revelation games'', in which
two players are given the choice to truthfully reveal private information before commencing
 two rounds of alternating negotiation. Revealing this information narrows the
search space of possible agreements and may lead to agreement more quickly, but may also
cause players to be exploited by others (see examples of such games in Figure~\ref{fig:ctfig}).
Revelation games combine two types of interactions that have been studied in the past in
the economics literature: Signaling games \cite{spence1974market}, in which players choose whether to convey
private information to each other, and bargaining \cite{osborne1994course}, in which players engage in multiple
negotiation rounds.

We were hopeful that, for revelation games, equilibrium-based agents will interact well with people since
behavioral economics work has shown that people often follow equilibrium strategies \cite{banks1994experimental}
when deciding whether to reveal private information to others. 
The question is whether this observation will be stronger than our previous observations reported above that people's behavior in bargaining settings does not adhere to equilibrium 
strategies. We formalized the setting as a Bayesian game and computed two types of perfect Bayesian equilibrium: a separating equilibrium where both players reveal their type, and a pooling equilibrium where none of 
the players reveal their types.    

We compared the equilibrium agents with people playing with
other people and with the Sigmoid Acceptance Learning Agent (SIGAL) that we developed \cite{peled2015study}.
The SIGAL agent used classical
machine learning techniques to predict how people make and respond to
offers during negotiation, how they reveal information and their
response to potential revelation actions by the agent. This model is
integrated into the agent's decision tree.  We conducted an
extensive empirical study spanning hundreds of human subjects. 

Results
show that the SIGAL agent was able to outperform people and the equilibrium agents. Furthermore, people outperformed the equilibrium agents.
It turned out that the negotiation part of the game was more important (with respect to the utility score) than the revelation part. 
The equilibrium agent made very selfish offers in the last round of the negotiations. Most of
these offers  were rejected. In the first round, it made offers that were highly beneficial
to people and most of these offers 
were accepted, but the small benefit it incurred in these proposals did not aid its performance.
The SIGAL agent, on the other hand, (i) learned to make offers that were beneficial to people while not
compromising its own benefit; and (ii) incrementally revealed
information to people in a way that increased its expected
performance. 
We were able to adjust SIGAL to new, similar settings that varied
rules and situational parameters of the game without the need to
accumulate new data. However, moving to a completely new setting requires a lot of work collecting data and adjusting the machine learning module to the new setting.

\subsection{Non-binding agreements}

\begin{figure}[t]
\centering
   \includegraphics[width=3.5cm]{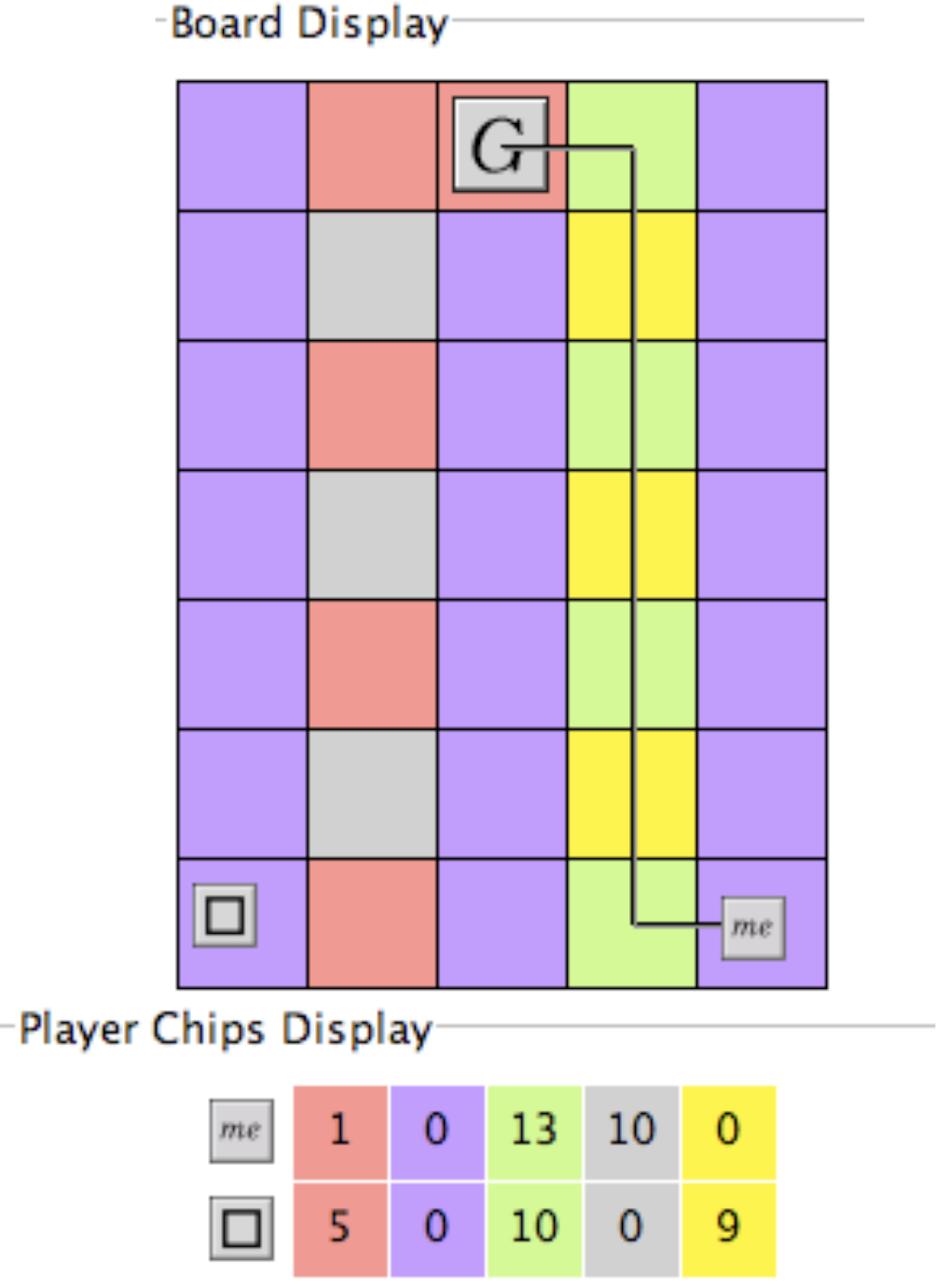}
\caption{An example of a CT Board for multiple negotiation games with unenforceable agreements.}
\label{fig:ddBoard}
\end{figure}

We also studied CT settings of two players in which both
participants needed to complete their individual tasks by reaching
agreements and exchanging resources, the number of negotiation rounds
were not fixed in advance, and the negotiation protocol was an
alternating offers protocol that allowed parties to choose the extent
to which they kept each of their agreements during the negotiation \cite{HaimGGK12}.
That is, there are three phases in each round of the game: negotiation, transfer and movement. 
The negotiation phase consisted of two rounds of alternating offers in which the players could reach an agreement on resource exchange.
After each phase of negotiations, the game moved to the ``transfer phase'' in which both players could 
transfer resources to each other. The transfer action was done simultaneously,
such that neither player could see what the other
player transferred until the end of the phase. A player could
choose to transfer more resources than it agreed to, or any subset
of the resources it agreed to, including not transferring any resources at
all. In the ``movement phase'' both players could move their
icons on the board one step towards the goal square, provided
they had the necessary resources. Then, the game moved to the next round, beginning again with negotiation phase.
The game ends when one of the players reaches his goal or does not move for two rounds (see an example of one such game in Figure~\ref{fig:ddBoard}).

The most important decision of a player in such settings is whether or not to keep the agreements. Another important decision is whether to accept an offer given by the other player.
In subgame perfect equilibrium, the players should not keep the agreements. Different equilibria may specify various strategies for the acceptance decision.
We ran preliminary experiments and observed that such strategies are not beneficial when the equilibrium agent interacts with people. Most of the time the agent was not able to reach its goal, yielding a low utility score.

Galit et al. \cite{HaimGGK12} present the Personality Adaptive Learning (\textsc{PAL}) agent for negotiating with people
from different cultures for the CT game where agreements are not enforceable.
The methodology was similar to that of SIGAL (Section~\ref{SIGAL-sec}), combining a decision-theoretic
model using a decision tree with classical machine learning techniques
to predict how people respond to offers, and the extent to which they
fulfill agreements.

\textsc{PAL} was 
evaluated empirically in the Colored Trails (CT) environment by
playing with people in three countries: Lebanon, the U.S., and Israel,
in which people are known to vary widely in their negotiation
behavior. The agent was able to outperform people in all three
countries.

\subsection{Contract Game}

\begin{figure}[t]
\centering
   \includegraphics[width=3.5cm]{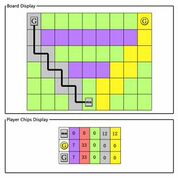}
\caption{An example of a CT Board for the Contract game.}
\label{fig:contract}
\end{figure}

We studied commitment strategies in a three-player CT game. The game is called Contract Game and is analogous
to a market setting in which participants need to reach agreements over
contracts and commit to or renege from contracts over time in order to succeed \cite{HaimGKA14}. The game comprises three players, two service providers and one
customer. The service providers compete to make repeated contract offers
to the customer consisting of resource exchanges in the game (see an example of one such game in Figure~\ref{fig:contract}). We formally
analyzed the game to compute subgame perfect equilibrium strategies
for the customer and service provider in the game that are based on
making contracts containing commitment offers. To evaluate
agents that use the equilibrium strategies, we conducted extensive empirical studies in three different countries, the U.S., Israel and China. We
ran several configurations in which two human participants played a single
agent participant in various role configurations in the game. Our results
showed that the computer agent using subgame Nash equilibrium strategies for the
customer role was able to outperform people playing the same role in all
three countries and obtained statistically significant, higher utility scores than the humans.
This was very surprising since it was the first EQ agent after trying many equilibrium agents that was able to achieve such results.

In particular, the customer agent made
significantly more commitment type proposals than people did, and requested significantly more resources from service providers than did people. 
It was quite surprising that people playing with it accepted these offers; in other settings (such as the revelation games) such unfair offers were rejected by people.
We hypothesize that the competition between the two service providers made such offers more acceptable.
In addition, while in the revelation games the EQ agent had only one opportunity to make an offer, in the contract game it could make offers several times (off the equilibrium path) which we believe also 
increased the acceptance rate. 
Also, the customer
agent reached  one of the goals in all its games and was able to reach the goal significantly more often than people. This is again quite surprising since at the beginning of the game the customer has enough resources to reach both goals.
Since reaching one of the goals is very beneficial to the customer it is difficult to understand why human players hadn't always reached the goal.

While the customer EQ  agent outperformed people, people outperformed the EQ agent when it played the role of one of the service providers.
We believe that this is mainly due to people playing the customer role not reaching the goal even when they have all the needed resources to do so.
To face this problem we then developed an agent termed SP-RAP which extended the EQ agent in the following two
ways to handle the uncertainty that characterizes human play in negotiation:
First, it employed a risk averse strategy using a convex utility function.
Second, it
reasoned about a possibly bounded rational customer (CS) player by assigning a positive
probability $p > 0$ for the customer player not reaching the goal. We assigned a
separate value for $p$ for each country by dividing the number of times the CS
player reached the goal by the total number of games played.
Consequently, SP-RAP outperformed people playing the SP role in all three countries.

\section{Argumentation Agent}

An automated agent can help a human when engaging in an argumentative dialog by utilizing its knowledge and computational advantage to provide arguments to him.
Argumentation was studied extensively using the well-established Argumentation Theory (see \cite{walton2009argumentation} for a summary).
Therefore, in the first step in the development of an automated agent that advised people in such settings, we considered
the abilities of Argumentation Theory to predict people's arguments. 
In \cite{rosenfeld2015providing} 
we presented extensive studies in three experimental settings, varying in complexity, which show the lack of predictive power of the existing Argumentation Theory. 
Second, we used Machine Learning (ML) techniques to provide a probability distribution over all known arguments given a partial deliberation. 
That is, our ML techniques provided the probability of each argument to be used next in a given dialog. Our model achieves 76\% accuracy when predicting people's top three argument choices given a partial deliberation.
Last, using the prediction model and the newly introduced heuristics of relevance, we designed and evaluated the Predictive and Relevance based Heuristic agent (PRH). 
Through an extensive human study, we showed that the PRH agent outperforms other agents that propose arguments based on Argumentation Theory, 
predicted arguments without heuristics or only the heuristics on both axes we examined: people's satisfaction from agents and people's use of the suggested arguments.

\section{Security Games}

The last several years have witnessed the successful application of
Bayesian Stackelberg games in allocating limited resources to
protect critical infrastructures. These interesting efforts have been led by Prof. Milind Tambe from USC.
The first application is the ARMOR system (Assistant for Randomized Monitoring
over Routes) that has been deployed at the Los
Angeles International Airport (LAX) since 2007 to randomize
checkpoints on the roadways entering the airport
and canine patrol routes within the airport terminals \cite{paruchuri2008playing,pita2008deployed}.
Other applications include IRIS, a game-theoretic scheduler for randomized
deployment of the US Federal Air Marshal Service
(FAMS) requiring significant scale-up in underlying algorithms, which
has been in use since 2009 \cite{tsai2009iris}; and
PROTECT, which requires further scale up, is deployed for
generating randomized patrol schedules for the US Coast
Guard in Boston, New York, Los Angeles and other ports
around the US \cite{an2011mixed,an2013deployed}. Furthermore, TRUSTS has been evaluated
for deterring fare evasion, suppressing urban crime
and counter-terrorism within the Los Angeles Metro System \cite{yin2012trusts,jiang2013game,delle2014game} and GUARDS was
earlier tested by the US Transportation Security Administration
(TSA) for security inside the airport \cite{pita2011guards}.

The evaluation of these systems could be quite limited. The only system that was truly evaluated in the field is TRUSTS. 
We were able to conduct controlled experiments of our game theoretic resource allocation algorithms. 
Before this project, the actual evaluation of the deployed security games applications in the field was a major open challenge. 
The reasons were twofold. First, previous applications focused on counter-terrorism, therefore controlled experiments against real 
adversaries in the field were not feasible. Second, the number of practical constraints related to real-world deployments limited the ability of researchers to conduct head-to-head comparisons

In TRUSTS we were able to address this challenge and run the largest scale evaluation of security games in the field in 
terms of duration and number of security officials deployed. We evaluated each component of the system (Fare Evasion, Counter Terrorism and Crime algorithms) 
by designing and running field experiments. In the context of fare evasion, we ran an extensive experiment, where we compared schedules 
generated using game theory against competing schedules comprised of a random scheduler, augmented with officers providing real-time knowledge 
of the current situation. Our results showed that our schedules led to statistically significant improvements over the competing schedules, 
despite the fact that the latter were improved with real-time knowledge. 

In addition, extensive human experiments in the lab were conducted \cite{pita2010robust,nguyen2013analyzing}. These experiments showed that incorporating bounded rational models of the adversaries to the 
Stackelberg games improves the performance of the defenders. These results were observed both when the role of adversaries was played by novices and when it was played by security experts.

\section{Persuasion Games}

A persuasion game involves two players: a sender who attempts to persuade another
agent (the receiver) to take a certain action \cite{glazer2006study}. Persuasion games are similar to both negotiation games and security games \cite{xu2015exploring}. 
They are similar to negotiation and argumentation games since one player tries to convince another player to do something, as in negotiations. They are also similar to security games in the asymmetry 
between the players: the sender and the defender are trying to influence the activities of the receiver and the attacker, respectively. So, it is interesting to check if equilibrium strategies will be 
beneficial in persuasion games. Furthermore,  the incorporation of a bounded rational model of the receiver will be beneficial to the sender as the incorporation of 
bounded rational models of the attacker was beneficial to the defender in security games.

We focused on information disclosure games with two-sided uncertainty \cite{azaria2014strategic,AzariaRKG11}. This is a special type of persuasion game in
which an agent tries to lead a person to take an action that is beneficial to the agent
 by providing him with truthful, but possibly partial, information relevant to the action
selection. We first computed the subgame
perfect  Bayesian Nash equilibrium of the game assuming the human receiver is fully rational.
We developed a sender agent that follows the equilibrium strategy (GTBA agent).

We also developed a machine learning-based model that effectively
predicts people's behavior in these games and we called it Linear weighted-Utility Quantal response
(LUQ). The model we provide assumes that people
use a subjective utility function which is a linear combination for all given attributes.
The model also assumes that while people use this function as a guideline, they do not
always choose the action with the greatest utility value, however, the higher an action's
utility value is, the more likely they are to choose that action. We integrated this model
into our persuasion model and built the LUQA agent.

We ran an extensive empirical study with people in two different games. In a multi-attribute road selection game with
two-sided uncertainty, the LUQA agent obtained significantly higher utility points than the GTBA agent.   
However, in the second game, the Sandwich game, there was no significant advantage
to the machine learning-based model, and using the game theory-based agent, GTBA,
which assumes that people maximize their expected monetary values is beneficial. 
We hypothesize that these different results are due to the nature of the domains. The monetary result plays an important role in the sandwich
game. This is because the game is played in an environment where a person's goal is to make a profit. However, in the road selection game the utility scores are associated with time.
Thus, it seems that maximizing expected monetary utility is easier for people than maximizing utility scores that are associated with time.

\section{Discussions}

The state-of-the-art agent, NegoChat, for multi-issue negotiations integrates methods from several disciplines: qualitative decision-making, machine learning and heuristics based on psychological theories. 
None of the equilibrium agents that  were developed were successful when interacting with people.
The reliance on heuristic and machine learning makes the transfer of NegoChat from one scenario to the other and from one culture to the other problematic.
This was evident recently when we tried to run experiments with NegoChat, which was developed based on data collected in Israel and Egypt. We had to spend a lot of time and effort until this transfer was possible.

Similarly, in most of the cases, the equilibrium agent was not successful in the CT game settings.
The only exception is the contract game. We believe that the success of the equilibrium agent in the contract game has to do with the specifics of the game: the competition between the two SPs.
In the contract game, it was extremely difficult to predict people's behavior, thus the success of the equilibrium agent is even more significant.

The same observations were seen in argumentation -- the argumentation theory-based agent was not very successful.
Therefore, in all these cases the development of new negotiation agents to new settings requires the collection of data  and the adjustment of the agent to the new settings.
Therefore, we strongly believe that the development of theoretical models for the design and implementation of agents that negotiate in multi-issue negotiation settings can be very useful.
However, this is still an open question.    

On the other hand, it seems that in security games the deployment of Stackelberg equilibrium 
is beneficial (possibly with the incorporation of a bounded rational model of the attacker) and similarly in persuasion games where using subgame perfect Baysian equilibrium is beneficial 
(possibly with the incorporation of a bounded rational model of the receiver).

We hypothesize that this is the case since in security games and persuasion games the interactions between the agent and the human is quite limited.
The attacker or the receiver needs to choose one action compared to many decision-making activities that are required from a human negotiator. 
Nevertheless, even in security games and to some extent in persuasion games it was shown that taking the limitations of the other player into consideration is beneficial.

\section{Acknowledgment}
This work was supported in part by ERC grant \#267523.

\bibliographystyle{eptcs}
\bibliography{tark}

%
%
\end{document}